%% file: iclr2027_conference.tex
\documentclass{article} 
\usepackage{iclr2027_conference,times}

\input{math_commands.tex}

\usepackage{hyperref}
\usepackage{url}
\usepackage{paralist}
\usepackage{booktabs}        
\usepackage{graphicx}        
\usepackage{amsmath,amssymb} 
\usepackage{subcaption}
\usepackage{makecell}
\usepackage{float}
\usepackage[svgnames,table]{xcolor}
\newcommand{\gain}[1]{\textcolor{green!55!black}{\scriptsize (+#1)}}
\newcommand{\loss}[1]{\textcolor{red!55!black}{\scriptsize (-#1)}}
\newcommand{\zero}[1]{\textcolor{black!55}{\scriptsize (#1)}}
\usepackage{multirow}
\usepackage{adjustbox}
\usepackage{hyperref}
\usepackage{pifont}
\newcommand{\cmark}{\textcolor{green!55!black}{\ding{51}}}
\newcommand{\xmark}{\textcolor{red!70!black}{\ding{55}}}

\title{GraphForge: Training Working Agents with Graph-Anchored Workspace Synthesis}

\author{\textbf{Qisheng Su$^{1,3}$, Hanchen Wang$^{2}$, Guanru Zhu$^{2}$, Huicheng Jiang$^{2}$, Qiuyinzhe Zhang$^{1}$,}\\ 
\textbf{Kou Shi$^{1}$,
Zhen Fang$^{1}$, Ziao Zhang$^{1}$, Qingnan Ren$^{1}$, Honglin Guo$^{2}$, Jiazheng Zhang$^{2}$,} \\
\textbf{Zehui Chen$^{1}$, Tao Gui$^{2}$, Feng Zhao$^{1}$} \\
\\
$^{1}$University of Science and Technology of China \quad
$^{2}$Fudan University \\
$^{3}$Shanghai Innovation Institute \\
\texttt{nicksu@mail.ustc.edu.cn} \quad 
\texttt{tgui@fudan.edu.cn} \quad
\texttt{fzhao956@ustc.edu.cn}
}

\iclrfinalcopy 
\begin{document}

\maketitle

\begin{abstract}
Working agents need to read diverse files, coordinate tools, and produce deliverables. Training such agents requires tasks built on many real files with verifiable results, but few pipelines exist to synthesize this kind of data. Existing pipelines either generate files with models, which lack realism and diversity, or build tasks on real files without task-specific verifiers, leaving result quality unchecked. We introduce GraphForge, an evidence-graph based framework that grounds both the task and its verification in real files. Starting from occupation-grounded seeds for controlled diversity, GraphForge assembles a workspace of real files for each seed and builds an evidence graph over their relations. Since the task statement and rubrics are both derived from this graph, task requirements are backed by the workspace files and each criterion is anchored to the files needed to verify it. An initial rollout further tests executability, and a revision agent repairs the task and rubrics against the original files before trajectories are collected. Fine-tuning Qwen3.6-27B on 2,169 GraphForge trajectories brings GDPVal to 1445.7 (+65.7) under OpenHands, and Workspace-Bench-Lite and SpreadsheetBench II to 63.7 (+7.7) and 24.0 (+13.7) under Claude Code. Rejection fine-tuning on the SFT model's own rollouts, with candidates selected by the evidence-anchored rubrics, yields further improvements on all three benchmarks, suggesting that the rubrics provide a useful selection signal. The data and models are available at \href{https://huggingface.co/collections/groundhogLLM/graphforge}{https://huggingface.co/collections/groundhogLLM/graphforge}.
\end{abstract}

\section{Introduction}
\label{sec:intro}
Large language model agents are moving from conversation to real work. Working agents such as OpenClaw \citep{openclaw} and Hermes-Agent \citep{hermesagent} act as persistent digital assistants, handling long-horizon tasks across file systems, databases, and terminal shells. Benchmarks such as Claw-Eval \citep{claweval}, GDPVal \citep{gdpval}, and Workspace-Bench \citep{workspacebench} evaluate these agents on realistic work tasks. Working agents read diverse files, coordinate tools, and produce deliverables that others can use. For agents more broadly, a common training approach is supervised fine-tuning on synthesized trajectories produced by a strong teacher model \citep{terminalworld,agentworld,facet}. Applying this approach to working agents requires tasks built on many real files, with verifiable results.

Two recent pipelines synthesize training data for working agents. EnvCraft \citep{envcraft} generates files with a model and checks each task with a Python script over the workspace state, which cannot read file contents and may overlook errors in document deliverables. NexForge \citep{nexforge} builds tasks on real files, but provides no task-specific rubrics or verifiers, so result quality cannot be systematically checked. It remains hard to construct diverse tasks on real files and to equip them with reliable verification rubrics.

To make progress on both fronts, we introduce GraphForge, an evidence-graph based framework for synthesizing working agent training data. Our design gives seeds and files separate roles. Tasks invented freely by a model tend to collapse toward frequent occupations and generic task types. Grounded seeds therefore fix the task direction, keeping diversity controllable, while the concrete task and its verification are derived from the files. Our seeds are drawn from O*NET occupations and their official work activities, and for each seed, an agent crawls real files to form a workspace, over which a model builds an evidence graph of cross-file relations. The graph is then compiled into task statements and rubrics, so task requirements are backed by the workspace files and each criterion is anchored to the files needed to verify it. An initial rollout further tests executability, and a revision agent repairs the task and rubrics against the original files before trajectories are collected.

We validate our framework by training Qwen3.6-27B on the synthesized data. Supervised fine-tuning on 2,169 trajectories from GraphForge brings GDPVal to 1445.7 (+65.7), Workspace-Bench-Lite to 63.7 (+7.7), and SpreadsheetBench~II to 24.0 (+13.7). The same data also improves Qwen3.6-35B-A3B, suggesting that GraphForge trajectories generalize across base models. Rejection fine-tuning on the SFT model's own rollouts on new queries disjoint from the SFT data yields further improvements on all three benchmarks, suggesting that the evidence-anchored rubrics provide a useful selection signal.

Our contributions are as follows.
\begin{compactitem}
    \item We propose GraphForge, a framework that grounds both the task and its verification in real files. Starting from occupation-grounded seeds, task statements and rubrics are compiled from an evidence graph over real files, and each task is validated through an initial rollout before trajectory collection.
    \item Training Qwen3.6-27B and Qwen3.6-35B-A3B on GraphForge data yields large gains on GDPVal, Workspace-Bench-Lite, and SpreadsheetBench~II, and our analysis suggests that rubric-guided selection adds signal beyond training on the model's own rollouts.
    \item We release the GraphForge dataset and the trained models to support future research on working agents.
\end{compactitem}

\begin{figure}[t]
\centering
\includegraphics[width=\textwidth]{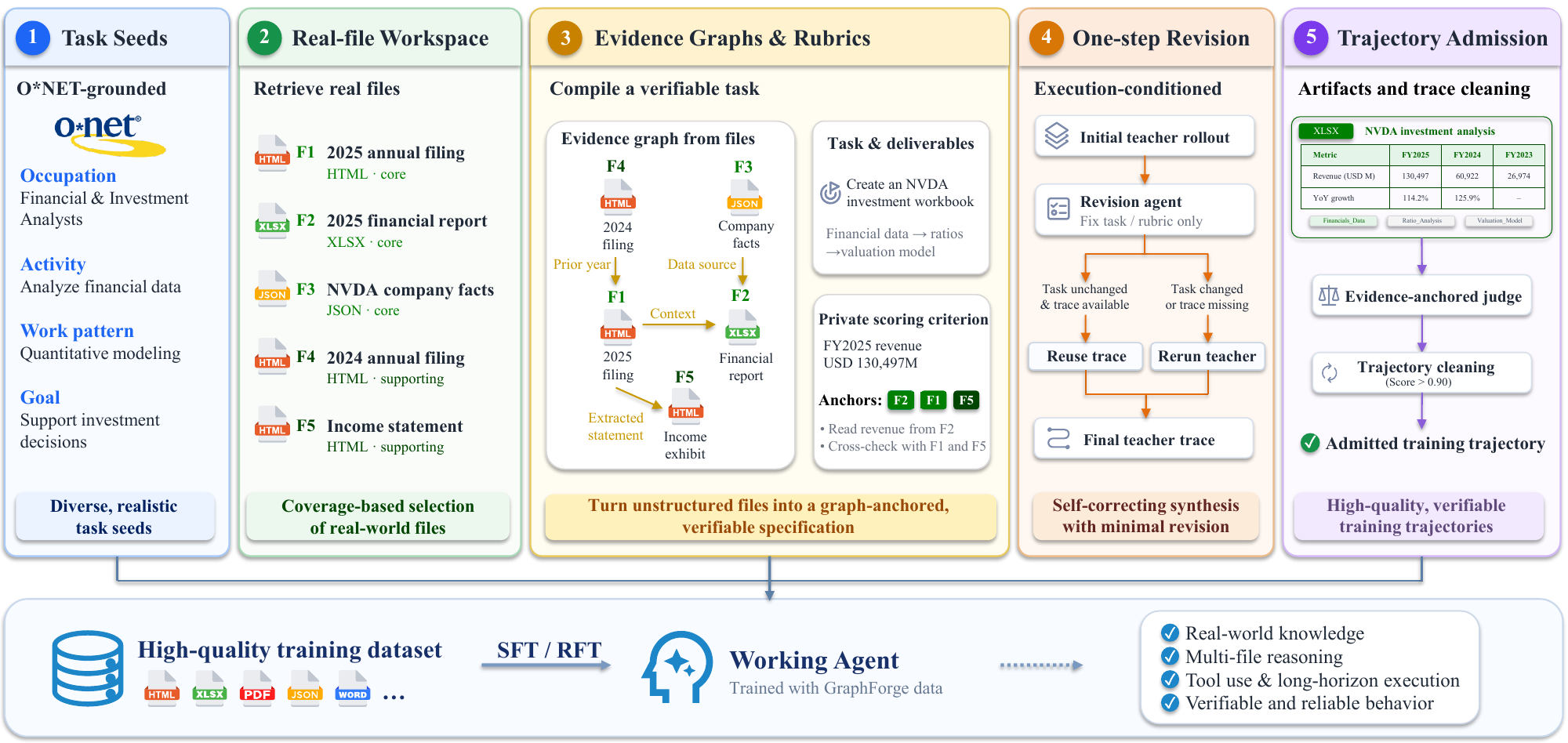}
\caption{\textbf{Overview of the GraphForge pipeline.} Starting from an O*NET-derived seed, an agent assembles a workspace of real files with hidden roles, and a model builds an evidence graph that is compiled into a task specification whose rubric criteria are anchored to graph nodes. An initial teacher rollout supports a one-step revision of the task specification, and the final trajectory is scored by an evidence-anchored judge before admission.}
\label{fig:pipeline}
\end{figure}

\section{Related Work}

\textbf{Agent task and environment synthesis.} Recent work synthesizes tasks and environments for agent training across several domains, including general tool use \citep{agentworld,awm,taskcraft}, computer use \citep{agentsynth}, software engineering \citep{swesmith,r2egym}, and terminal operation \citep{skillsynth,terminalworld,cliuniverse,facet,terminaltraj,openthoughtsagent}. In these domains, task outcomes can be verified programmatically. The most relevant pipelines to our work are EnvCraft \citep{envcraft} and NexForge \citep{nexforge}, discussed in Section~\ref{sec:intro}. GraphForge addresses their shared gap by compiling an evidence graph over real crawled files into rubrics with evidence anchors, so that trajectory selection and final evaluation are both grounded in source files.

\textbf{Evaluation of working agents.} A number of recent benchmarks evaluate agents on realistic work tasks, each with a different emphasis. GDPVal \citep{gdpval} covers 1,320 tasks across 44 occupations, grades deliverables with expert-written rubrics, and reports Elo ratings from pairwise comparisons against human work. Workspace-Bench \citep{workspacebench} places agents in realistic workspaces with tens of thousands of files and evaluates cross-file dependency reasoning with fine-grained rubrics. APEX-Agents \citep{apexagents} focuses on professional services, with tasks created by investment banking analysts, management consultants, and corporate lawyers inside data-rich simulated worlds. Claw-Eval \citep{claweval} grades agents with trajectory-aware evidence, recording execution traces, audit logs, and environment snapshots to score fine-grained rubric items along completion, safety, and robustness. Agora\citep{agora} tests archive-grounded reasoning, requiring agents to locate sparse evidence across large collections of authentic workplace documents and reconcile inconsistencies across them. SpreadsheetBench~II \citep{spreadsheetbench2} evaluates spreadsheet agents on end-to-end business workflows across generation, debugging, and visualization, with expert-annotated tasks over multi-sheet workbooks built from authentic business data. We evaluate our trained models on GDPVal, Workspace-Bench, and SpreadsheetBench~II, following the official protocol of each benchmark.

\begin{table}[t]
\caption{\textbf{Comparison of training-data synthesis pipelines for agents.} The upper block lists general-domain pipelines, the middle block lists working-agent pipelines, and the bottom row shows our pipeline.}
\label{tab:datasets}
\centering
\small
\resizebox{\columnwidth}{!}{%
\begin{tabular}{@{}lllllc@{}}
\toprule
\textbf{Dataset} & \textbf{Domain} & \textbf{Task environment} & \textbf{Verification} & \textbf{Scale} & \textbf{Open data} \\
\midrule
\rowcolor{MistyRose!40}
\multicolumn{6}{@{}l}{\textit{\textbf{General-Domain Pipelines}}} \\
AgentSynth & computer use & desktop VM & per-step execution check & 6K tasks & \cmark \\
TaskCraft & tool use & web and document tools & golden answer & 36K tasks & \cmark \\
SWE-smith & software eng. & code repositories & fail-to-pass tests & 50K tasks & \cmark \\
CLI-Universe & terminal & docker environments & fail-to-pass tests & 6K trajs & \xmark \\
\midrule
\rowcolor{Honeydew}
\multicolumn{6}{@{}l}{\textit{\textbf{Working-Agent Pipelines}}} \\
EnvCraft & working & synthesized workspaces & state-check scripts & 20K tasks & \cmark \\
NexForge\textsuperscript{1} & working & real files & none & 5.6K tasks & \xmark \\
\midrule
\rowcolor{AliceBlue}
\multicolumn{6}{@{}l}{\textit{\textbf{Our Pipeline}}} \\
\textbf{GraphForge} & \textbf{working} & \textbf{real files} & \textbf{evidence-anchored agent judge} & \textbf{2.1K tasks} & \textbf{\cmark} \\
\bottomrule
\end{tabular}%
}
{\raggedright\footnotesize \textsuperscript{1}\,NexForge releases trained models but not the synthesized tasks or trajectories.\par}
\end{table}

\section{Method}

GraphForge turns occupational task seeds into verifiable training trajectories through five stages, from seed construction to trajectory admission. \textbf{Our design gives seeds and files separate roles.} Seeds fix the occupational direction, so diversity is controlled at seed selection and rebalanced after workspace materialization. The concrete task and its verification are derived \textbf{only after} a real workspace has been instantiated, so task requirements and evaluation signals are both grounded in source evidence.

\subsection{O*NET-grounded task-form seeds}

Our seeds come from the O*NET database, which provides occupations, their task statements, and a controlled vocabulary of Detailed Work Activities (DWAs) with an official task-to-DWA mapping \citep{onet}. As not all tasks are digitally executable, we keep only those annotated DIGITAL by AI4Work \citep{ai4work}. Each retained task is mapped to its DWA through the official relation, and the DWA serves as our controlled task type. After filtering, 246 occupations across 16 sectors and 43 sub-sectors remain, covering 891 DWA task types and 3,419 valid occupation-task-type pairs.

Each seed is a tuple
\[
s_i=(o_i,a_i,w_i,p_i,u_i,e_i^{\mathrm{occ}}),
\]
where $o_i$ is an occupation, $a_i$ a DWA task type, $w_i$ an occupation-specific work demand, $p_i$ the dominant execution pattern, $u_i$ the expected input file family, and $e_i^{\mathrm{occ}}$ the retrieved occupational evidence. For each candidate pair $(o_i,a_i)$, we retrieve professional passages and keep only work demands $w_i$ directly supported by the cited evidence; unsupported demands are dropped rather than filled to a quota. A second step assigns each demand a dominant execution pattern $p_i$ from a vocabulary of 16, from quantitative modeling and reconciliation to policy design and artifact revision.

To avoid concentrating the corpus on frequent occupations or generic analysis tasks, we select seeds by marginal coverage over the dimensions $(o,a,p,u)$. Let $\mathcal{D}$ denote these dimensions and $n_d(v)$ the number of already selected seeds with value $v$ in dimension $d$. The gain of a candidate $s$ is
\[
\Delta(s\mid\mathcal{S})=\sum_{d\in\mathcal{D}}\frac{1}{1+n_d(v_d(s))}.
\]
We greedily pick the candidate with the highest gain, breaking ties by a stable task-ID hash. After files are downloaded and validated, the same rule is applied again to the actual input and output families. Balanced subsets add equal quotas over $p$, sector round-robin over $o$, and a cap on repeated normalized demands $w$.

\subsection{Real-file workspace construction}

\textbf{A seed $s_i$ is case-neutral.} Its occupation $o_i$, work demand $w_i$, and execution pattern $p_i$ specify who does the work, what demand is addressed, and how it is mainly carried out, but the seed names no company, event, dataset, or result. A search agent instantiates $s_i$ by finding a coherent public case and retrieving the files needed to do the work, forming a workspace $W_i=\{f_{i1},\dots,f_{im}\}$. Files are downloaded in native formats, parsed with format-specific tools, and exact duplicates and invalid files are removed.

Each retained file $f\in W_i$ gets a hidden role $\rho(f)\in\{\text{core},\text{supporting},\text{confuser},\text{ambient}\}$. Core files drive the main computation or decision, supporting files provide policy or context, confusers are plausible but inapplicable alternatives, and ambient files add realistic redundancy. These roles guide assembly and are never shown to the working agent. Files may span organizations, mixing related public evidence with same-domain distractors as long as the task stays coherent and answerable.

\subsection{Evidence graphs and verifiable rubrics}

Given the workspace $W_i$, GraphForge builds an evidence graph $G_i=(V_i,E_i)$. A node $v\in V_i$ records a source file, the fact or field it provides, and its role in the task. An edge $e\in E_i$ records a cross-file dependency needed to interpret, compare, reconcile, or derive information. The graph is not ground truth but an intermediate representation whose claims \textbf{must remain recoverable from the original files}.

The graph is compiled into a task specification
\[
\mathcal{C}_i=(q_i,\mathcal{D}_i,\mathcal{R}_i^{+},\mathcal{R}_i^{-};G_i),
\]
containing a natural task statement $q_i$, deliverable requirements $\mathcal{D}_i$, positive criteria $\mathcal{R}_i^{+}$, and negative penalty criteria $\mathcal{R}_i^{-}$. Each positive criterion
\[
c_k=(d_k,r_k,z_k,w_k,A_k,\phi_k),\qquad A_k\subseteq V_i,
\]
specifies the target deliverable $d_k$, the requirement $r_k$, the expected value or computation $z_k$, a weight $w_k$, evidence anchors $A_k$, and a verification procedure $\phi_k$. Negative criteria describe concrete prohibited outcomes and are penalized only when the violation is directly evidenced.

This design separates execution from verification. The working agent sees only $q_i$ and $W_i$, not node IDs or hidden roles $\rho(f)$. The judge receives the anchors $A_k$ and verification instructions $\phi_k$, telling it which files and deliverable parts to inspect. \textbf{Anchors thus guide both rubric generation and judging} without leaking a solution procedure into the task statement.

\subsection{Execution-conditioned one-step revision}

Static inspection cannot catch every ambiguity in a long-horizon task. We therefore run each initial task specification $\mathcal{C}_i^0$ once with a strong teacher model $\pi_T$, producing an initial trajectory $\tau_i^0$ and its deliverables. A revision agent then receives the original files $W_i$, the evidence graph $G_i$, the full task and rubrics, and this execution, and checks whether the task is natural and executable, whether required quantities are supported, whether every criterion is correctly anchored, and whether the verification instructions suffice to inspect the artifacts.

The revision agent returns only the components that need to change. The compiler keeps all untouched fields, validates references and schema constraints, and emits a revised specification $\mathcal{C}_i^1$. We rerun the teacher only when the task statement changes or the initial trajectory is missing. If only rubric bindings change, the initial execution is reused and judged against the revised specification. Formally, the trajectory retained for task $i$ is
\[
\tau_i=\begin{cases}
\tau_i^0, & H(q_i^1)=H(q_i^0)\ \text{and}\ \tau_i^0\ \text{exists},\\[2pt]
\pi_T(W_i,q_i^1), & \text{otherwise},
\end{cases}
\]
where $H(\cdot)$ denotes the normalized hash of the task statement. The first branch applies exactly when the revision leaves the task statement unchanged and a reusable initial trajectory is available.

\subsection{Artifact-level admission and trajectory cleaning}

A trajectory $\tau_i$ is admitted only after its promised deliverables are materialized. Deterministic checks verify required filenames, readable formats, required sheets and formulas in spreadsheets, and task-specific structural constraints. The agent judge then scores every criterion while consulting the referenced source files and produced artifacts. Each positive criterion contributes its weight times the fraction of the requirement met, and each negative criterion a penalty proportional to the evidenced degree of violation:
\[
Q_i(\tau)=
\frac{
\sum_{k\in\mathcal{R}_i^{+}} w^{+}_{ik}\, a_{ik}(\tau)
-
\sum_{j\in\mathcal{R}_i^{-}} \lambda_{ij}\, v_{ij}(\tau)
}{
\sum_{k\in\mathcal{R}_i^{+}} w^{+}_{ik}
},
\]
where $w^{+}_{ik}>0$ is the weight of a positive criterion, $a_{ik}\in[0,1]$ the fraction of the requirement met, $\lambda_{ij}>0$ the penalty strength of a negative criterion, and $v_{ij}\in[0,1]$ the evidenced degree of violation, with $v_{ij}=0$ when no violation is found. At this admission stage, violations are binary, so $v_{ij}\in\{0,1\}$. Scores are used raw and may fall below zero. We further discard trajectories with degenerate tool-use behavior, such as repeated non-polling calls, excessive tool use, high tool-failure rates, and repeated truncation.

\section{Main Results}
\label{sec:results}

We train Qwen3.6-27B and Qwen3.6-35B-A3B on GraphForge data and evaluate the resulting models on GDPVal-AA \citep{gdpval}, the 220-task gold subset of the full GDPVal benchmark, Workspace-Bench-Lite \citep{workspacebench}, and SpreadsheetBench~II \citep{spreadsheetbench2}. The first stage is SFT on admitted teacher trajectories. The second stage is RFT on rubric-selected trajectories generated by the SFT model itself.

\subsection{Training data}
\label{sec:data}

\begin{figure}[t]
\centering
\begin{subfigure}[c]{0.60\linewidth}
\centering
\includegraphics[width=\linewidth]{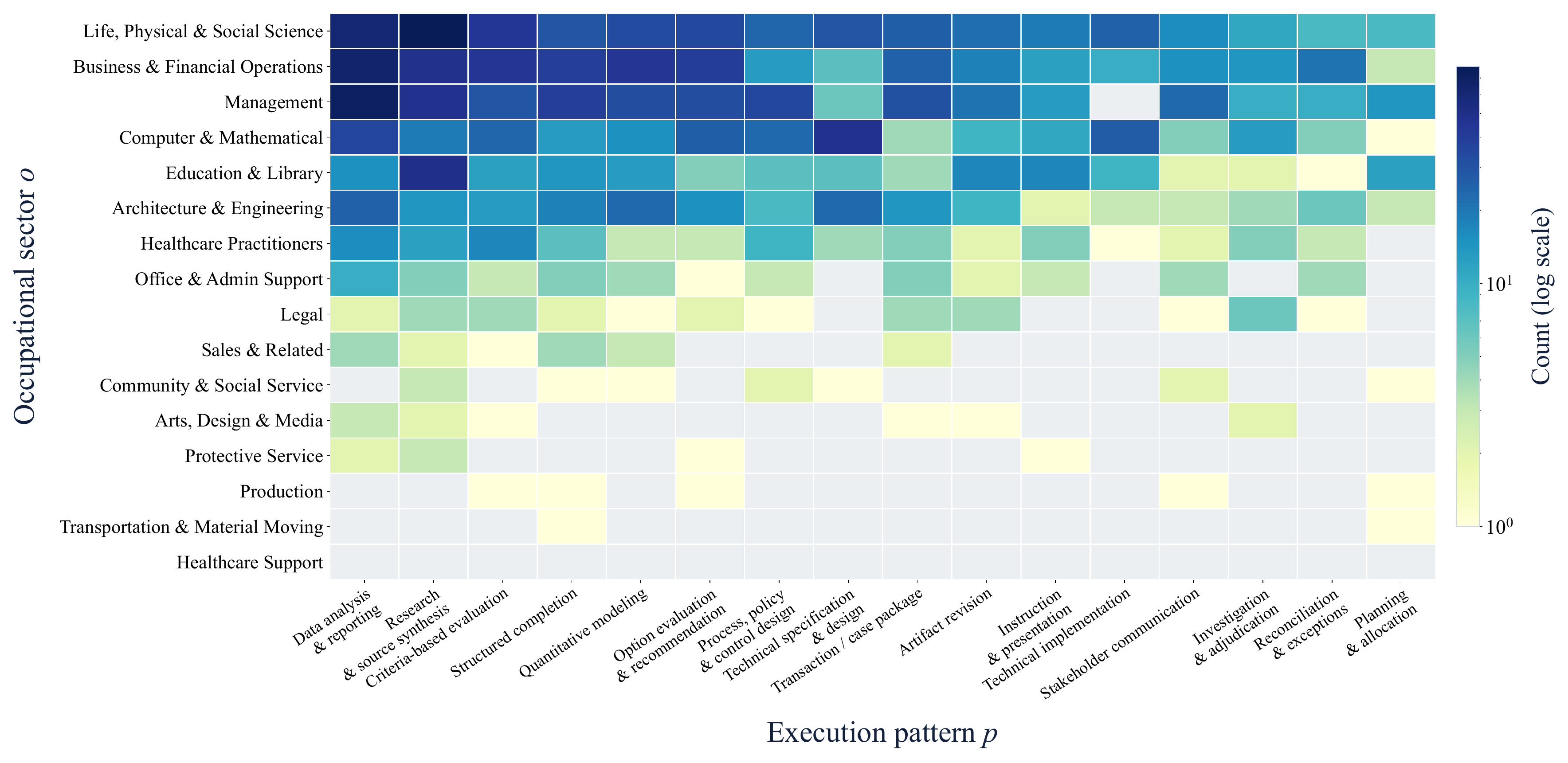}
\caption{Joint coverage of occupational sectors and execution patterns. Cell color gives the number of trajectories on a log scale, and gray cells are unobserved combinations.}
\label{fig:joint}
\end{subfigure}
\hfill
\begin{subfigure}[c]{0.38\linewidth}
\centering
\includegraphics[width=\linewidth]{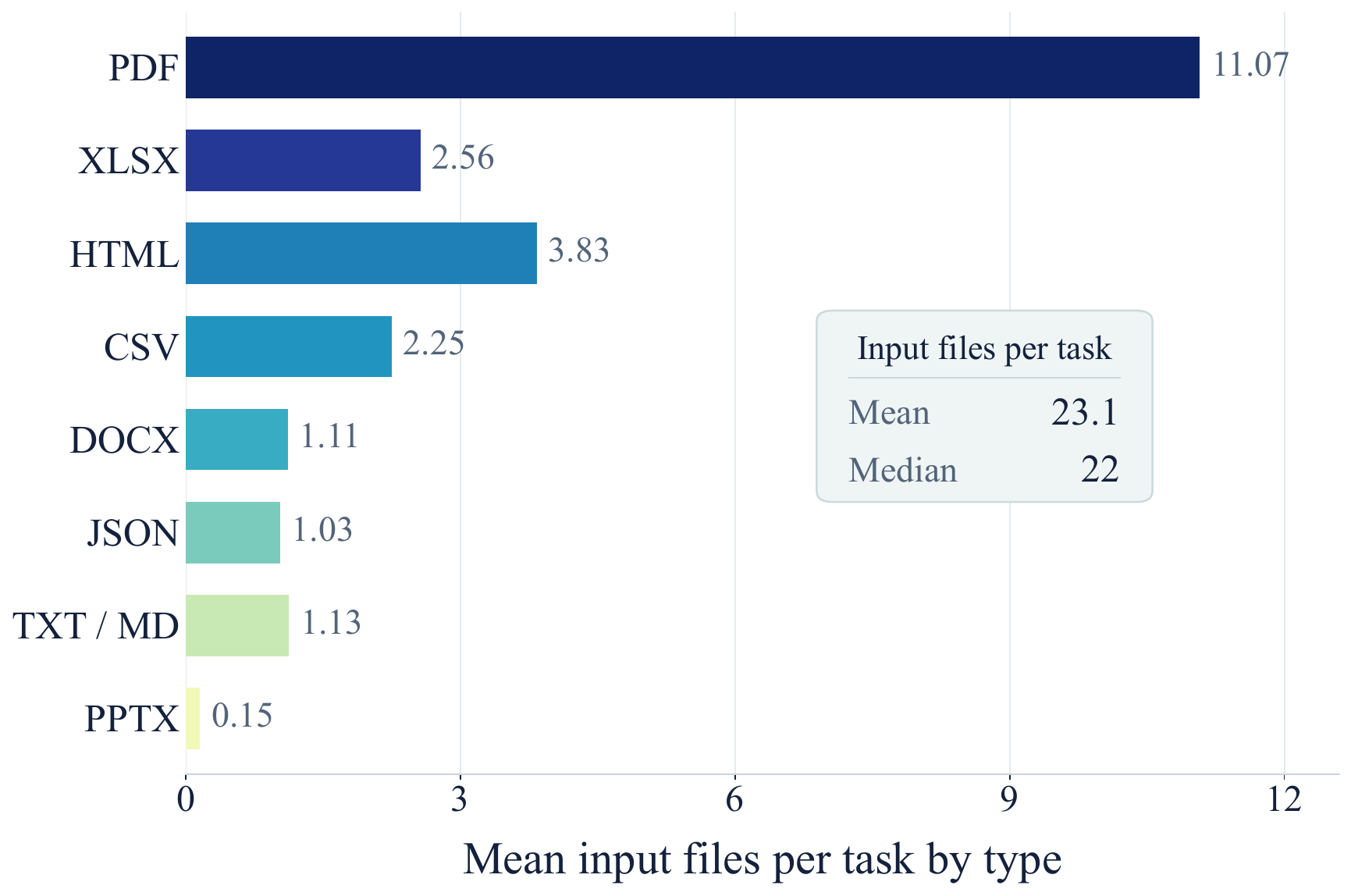}
\caption{Materialized input file families. One task can use several file families.}
\label{fig:input}
\end{subfigure}
\caption{Diversity of the SFT corpus across occupational sectors, execution patterns, and input file families.}
\label{fig:diversity}
\end{figure}

The SFT corpus contains 2{,}169 admitted trajectories with $Q_i(\tau_i) > 0.90$, selected from 3{,}638 materialized tasks through the construction funnel in Table~\ref{tab:audit-summary}(b). The corpus covers 466 distinct O*NET task types, 15 of the 16 occupational sectors, and all 16 execution patterns. \textbf{Tasks invented freely by a model tend to collapse toward frequent occupations and generic task types.} Our seeds fix the occupation, task type, execution pattern, and input family before any file is retrieved, and coverage-based selection keeps the corpus broad. Figure~\ref{fig:joint} shows the joint coverage of sectors and patterns, with 164 of the 256 sector--pattern combinations realized. The distribution is not uniform. Data analysis and reporting accounts for 14.2\%, research and source synthesis for 13.2\%, and no other pattern exceeds 9\%. This shape comes from occupational demand and admission filtering. Figure~\ref{fig:input} shows the materialized input file families. PDF appears in 96.7\% of the trajectories, and most tasks draw on several file families.

\begin{figure}[t]
\centering
\includegraphics[width=0.9\linewidth]{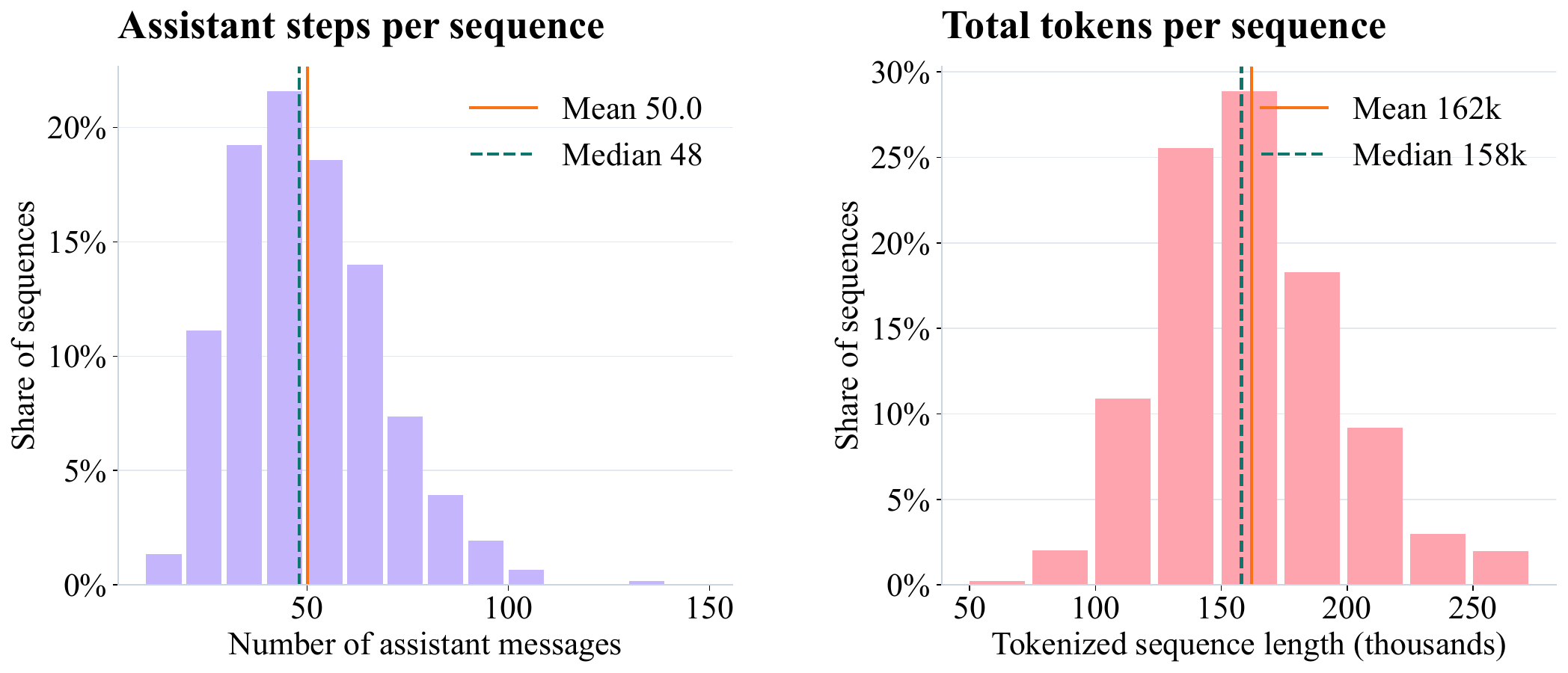}
\caption{Distribution of assistant steps and total tokenized length in the 2{,}169-example SFT corpus.}
\label{fig:length}
\end{figure}

Figure~\ref{fig:length} reports trajectory length. A trajectory contains 50.0 assistant steps on average (median 48, 95th percentile 82) and 162.0k tokens on average (median 158.0k, 95th percentile 224.8k). 28 sequences (1.3\%) reach the 262{,}144-token training ceiling.

\subsection{Training and evaluation setup}
\label{sec:setup}

\textbf{Training.} We use \textbf{GLM-5.2} for all components of the GraphForge pipeline, including the workspace construction agent, the evidence graph and task specification generation, the revision agent, the teacher rollouts, and the evidence-anchored judge. SFT trains on the admitted teacher trajectories, with the same corpus and optimization recipe for the 27B and 35B variants. Full configurations are in Appendix~\ref{app:training-config}. For RFT data selection, we sample $K=4$ rollouts per query from the SFT model on 2{,}000 queries, 125 per execution pattern. The evidence-anchored judge scores the four candidates of one query jointly against the rubrics of the task specification. We keep the highest-scoring valid trajectory when its score exceeds 0.95, apply the behavior filter, and drop queries where all candidates fail or the trajectory is overlength, giving 462 trajectories. The three RFT arms in Table~\ref{tab:rft} share the same 462 query IDs, candidate pools, and optimization budget, and differ only in the selection rule.

\textbf{Evaluation.} We evaluate on three working agent benchmarks, GDPVal-AA \citep{gdpval}, Workspace-Bench-Lite \citep{workspacebench} and SpreadsheetBench~II \citep{spreadsheetbench2}. Every model runs under pass@1 with a fixed workspace interface. Each scaffold (OpenHands, Codex, Claude Code) uses a fixed configuration, and comparisons between models are always made within the same scaffold. A missing or invalid deliverable, or a failure of the model to complete the task counts as a loss. Model-side timeouts and infrastructure failures are retried.

For GDPVal-AA we maintain an internal Elo pool. Each (model, scaffold) pair is a separate node, and we fit a Bradley--Terry model over the connected comparison graph~\citep{bradley1952rank,chiang2024chatbot}. Cross-scaffold bridge comparisons place the OpenHands and Codex nodes on a common scale. A tie contributes one half-win and one half-loss. Each node $i$ has a strength parameter $\theta_i$, and
\begin{equation}
\Pr(i \succ j) = \sigma(\theta_i - \theta_j), \qquad \mathrm{Elo}_i = 1667 + \frac{400}{\ln 10}\,(\theta_i - \theta_{\mathrm{anchor}}) .
\end{equation}
 The model is invariant to additive shifts of the strengths, so we anchor the scale by fixing the Elo of GLM-5.3 (OpenHands) to 1667. We report scores on the conventional Elo scale~\citep{elo1978rating,boubdir2023elo}, in which a 400-point gap corresponds to ten-to-one odds. The factor $400/\ln 10$ converts the fitted strengths to this scale.

\subsection{Overall comparison}
\label{sec:main}

\begin{table}[t]
\caption{\textbf{Overall comparison on working agent benchmarks.} GDPVal reports Elo under the OpenHands and Codex scaffolds, with each (model, scaffold) pair fitted as a separate node and the scale anchored at GLM-5.3 (OpenHands) = 1667. Workspace-Bench-Lite reports micro scores and SpreadsheetBench~II reports execution accuracy, both under the Claude Code and Codex scaffolds. GDPVal bootstrap confidence intervals are reported in Appendix~\ref{app:gdpval-elo-ci}.}
\label{tab:overall}
\centering
\resizebox{\columnwidth}{!}{%
\begin{tabular}{lcccccc}
\toprule
\multirow{2}{*}{\textbf{Model}} & \multicolumn{2}{c}{\textbf{GDPVal}} & \multicolumn{2}{c}{\textbf{Workspace-Bench-Lite}} & \multicolumn{2}{c}{\textbf{SpreadsheetBench~II}} \\
\cmidrule(lr){2-3} \cmidrule(lr){4-5} \cmidrule(lr){6-7}
& \textbf{OpenHands} & \textbf{Codex} & \textbf{Claude Code} & \textbf{Codex} & \textbf{Claude Code} & \textbf{Codex}\\
\midrule
\rowcolor{MistyRose!40}
\multicolumn{7}{l}{\textit{\textbf{Frontier Models}}} \\
Claude Opus 5   & 1774.1 & 1753.1 & 70.1 & 68.9 & 33.6 & ---  \\
GPT-5.6-sol     & 1687.1 & 1710.8 & ---  & 60.5 & ---  & 32.7 \\
Qwen3.8-Max     & 1719.0 & 1771.0 & 67.4 & 66.6 & 34.9 & 34.9 \\
GLM-5.3         & 1667.0 & 1543.7 & 67.7 & 61.4 & 32.1 & 31.5  \\
Kimi-K3         & 1615.5 & 1664.4 & 65.8 & 60.6 & 35.8 & 37.7 \\
DeepSeek-V4-Pro     & 1531.5 & 1576.7 & 58.1 & 57.9 & 29.3 & 35.5 \\
\midrule
\rowcolor{Honeydew}
\multicolumn{7}{l}{\textit{\textbf{Open-Weight Baseline}}} \\
Nex-N2-Mini-35B     & 1288.8 & 1342.3 & 33.1 & 31.6 & 6.5 & 10.3 \\
\midrule
\rowcolor{AliceBlue}
\multicolumn{7}{l}{\textit{\textbf{Our Models}}} \\
Qwen3.6-35B-A3B & 1260.6 & 1283.0 & 55.9 & 53.4 & 2.8  & 4.7 \\
\rowcolor{AliceBlue!15}
\textbf{\quad + SFT (GraphForge)}
  & \makecell{\textbf{1362.3}\\[-2pt] \gain{101.7}}
  & \makecell{\textbf{1384.4}\\[-2pt] \gain{101.4}}
  & \makecell{\textbf{59.7}\\[-2pt] \gain{3.8}}
  & \makecell{\textbf{60.0}\\[-2pt] \gain{6.6}}
  & \makecell{\textbf{19.3}\\[-2pt] \gain{16.5}}
  & \makecell{\textbf{18.7}\\[-2pt] \gain{14.0}} \\
Qwen3.6-27B     & 1380.0 & 1364.0 & 56.0 & 61.4 & 10.3 & 15.6 \\
\rowcolor{AliceBlue!15}
\textbf{\quad + SFT (GraphForge)}
  & \makecell{\textbf{1445.7}\\[-2pt] \gain{65.7}}
  & \makecell{\textbf{1427.4}\\[-2pt] \gain{63.4}}
  & \makecell{\textbf{63.7}\\[-2pt] \gain{7.7}}
  & \makecell{\textbf{65.2}\\[-2pt] \gain{3.8}}
  & \makecell{\textbf{24.0}\\[-2pt] \gain{13.7}}
  & \makecell{\textbf{24.6}\\[-2pt] \gain{9.0}} \\
\bottomrule
\end{tabular}%
}
\end{table}

Table~\ref{tab:overall} compares GraphForge with frontier models and with Nex-N2-Mini-35B, the open-weight model released with NexForge \citep{nexforge}. Supervised training on GraphForge data produces large gains on all three benchmarks. On GDPVal, SFT improves the 35B base model by 101.7 Elo on OpenHands and 101.4 Elo on Codex. The 27B SFT model reaches 1445.7 Elo on OpenHands and 1427.4 Elo on Codex, improving over its base model by 65.7 and 63.4 points. On Workspace-Bench-Lite, SFT improves the two base models by up to 6.6 and 7.7 points. On SpreadsheetBench~II, the gains reach 16.5 and 13.7 points.

The gain also transfers across agent scaffolds. All GraphForge trajectories are rolled out with the Codex scaffold, while the evaluation covers OpenHands and Codex on GDPVal and Claude Code and Codex on Workspace-Bench-Lite and SpreadsheetBench~II. The SFT model improves over the base model under every scaffold. This suggests that the corpus teaches working skills that transfer across scaffolds, rather than habits tied to the rollout scaffold.

\subsection{Rubric-guided rejection fine-tuning}
\label{sec:rft}

\begin{table}[t]
\caption{\textbf{RFT ablation on top of the SFT model.} Parentheses report the change from the SFT model. GDPVal values are SFT-anchored Elo from direct paired comparisons with SFT. GDPVal bootstrap confidence intervals are reported in Appendix~\ref{app:rft-elo-ci}.}
\label{tab:rft}
\centering
\resizebox{\columnwidth}{!}{%
\begin{tabular}{@{}lcccccc@{}}
\toprule
\multirow{2}{*}{\textbf{Model}} & \multicolumn{2}{c}{\textbf{GDPVal}} & \multicolumn{2}{c}{\textbf{Workspace-Bench-Lite}} & \multicolumn{2}{c}{\textbf{SpreadsheetBench~II}} \\
\cmidrule(lr){2-3} \cmidrule(lr){4-5} \cmidrule(lr){6-7}
& OpenHands & Codex & Claude Code & Codex & Claude Code & Codex \\
\midrule
\rowcolor{AliceBlue}
\multicolumn{7}{@{}l}{\textit{\textbf{Our Models}}} \\
Qwen3.6-35B-A3B (reference) & 1260.6 & 1283.0 & 55.9 & 53.4 & 2.8 & 4.7 \\
\rowcolor{AliceBlue!15}
\textbf{35B SFT (GraphForge)}
  & \textbf{1362.3}
  & \textbf{1384.4}
  & \textbf{59.7}
  & \textbf{60.0}
  & \textbf{19.3}
  & \textbf{18.7} \\
\midrule
\rowcolor{MistyRose!40}
\multicolumn{7}{@{}l}{\textit{\textbf{RFT Variants}}} \\
\textbf{\quad + RFT}
  & \makecell{1369.5\\[-2pt] \gain{7.2}}
  & \makecell{1395.4\\[-2pt] \gain{11.0}}
  & \makecell{\textbf{63.7}\\[-2pt] \gain{4.0}}
  & \makecell{\textbf{64.0}\\[-2pt] \gain{4.0}}
  & \makecell{\textbf{20.3}\\[-2pt] \gain{1.0}}
  & \makecell{\textbf{19.6}\\[-2pt] \gain{0.9}} \\
\textbf{\quad + RFT (unanchored)}
  & \makecell{1396.4\\[-2pt] \gain{34.1}}
  & \makecell{1409.7\\[-2pt] \gain{25.3}}
  & \makecell{\textbf{62.2}\\[-2pt] \gain{2.5}}
  & \makecell{\textbf{61.7}\\[-2pt] \gain{1.7}}
  & \makecell{\textbf{17.5}\\[-2pt] \loss{1.8}}
  & \makecell{\textbf{18.7}\\[-2pt] \zero{0.0}} \\
\textbf{\quad + RFT (random-of-4)}
  & \makecell{1353.3\\[-2pt] \loss{9.0}}
  & \makecell{1374.9\\[-2pt] \loss{9.5}}
  & \makecell{\textbf{60.8}\\[-2pt] \gain{1.1}}
  & \makecell{\textbf{62.8}\\[-2pt] \gain{2.8}}
  & \makecell{\textbf{19.0}\\[-2pt] \loss{0.3}}
  & \makecell{\textbf{15.6}\\[-2pt] \loss{3.1}} \\
\bottomrule
\end{tabular}%
}
\end{table}

We compare three offline rejection fine-tuning arms initialized from the same SFT checkpoint. From 2{,}000 newly synthesized queries, we sample up to four trajectories per query with the SFT model, forming a shared candidate pool. The anchored arm keeps, for each query, the rubric-best trajectory when its judge score exceeds 0.95. After validity and behavior filtering, 462 queries remain, each contributing one trajectory. The other two arms reuse the same 462 queries and their candidate pools. The unanchored arm ranks the candidates with a judge that sees the rubric text but not the explicit evidence anchors and verification instructions. The random-of-4 arm selects uniformly at random from the eligible candidates. All arms share the candidate eligibility rules, the 462 training examples, and the optimization budget, and each arm branches independently from the same checkpoint. This matched design isolates within-query trajectory selection rather than the full task-admission pipeline.

For scoring, each arm is compared directly against the SFT model under the same scaffold on the same tasks. We convert the resulting win rate into an Elo difference and add it to the frozen main-table SFT score, so all arms are reported on the same scale as Table~\ref{tab:overall}. These paired comparisons are independent of the joint pool used for the main table.

Table~\ref{tab:rft} reports the results. On Workspace-Bench-Lite and SpreadsheetBench~II, the ordering follows the design intent. Anchored selection gives the largest gains over SFT, unanchored selection gives smaller or negative gains, and random selection is the weakest overall. On GDPVal, anchored RFT improves over SFT (+7.2 and +11.0 Elo) and random selection degrades performance (-9.0 and -9.5), while the unanchored arm attains higher point estimates (+34.1 and +25.3). However, none of these GDPVal differences is statistically resolved at this scale. GDPVal Elo differences at 220 tasks are therefore noisy rather than decisive. We read the results as follows. Selection quality matters across benchmarks, since random selection is consistently the weakest arm. The advantage of evidence anchoring is reflected on Workspace-Bench-Lite and SpreadsheetBench~II, while GDPVal Elo is too noisy to separate the two judge variants. RFT is compatible with continued improvement and does not damage the SFT model.

\subsection{Ablation Studies}
\label{sec:ablation}

\textbf{Contamination and transfer.}
We audit all 2,150 training workspaces against the 220 GDPVal tasks at three levels of granularity (Table~\ref{tab:audit-summary}a), since GDPVal draws its tasks from the same O*NET taxonomy as our seeds and poses the highest overlap risk. At the file level, none of the 39,201 training files coincides with any of the 260 GDPVal files. At the text level, the top-20 most similar 13-gram pairs between the two corpora contain no substantive shared content. At the occupation level, only 13 of the 44 GDPVal occupations are covered by our training taxonomy, so most evaluation tasks are occupation-disjoint from the training data.

To test whether the SFT gain is concentrated near covered content, we split GDPVal tasks by occupation coverage and compare Base vs.\ SFT win rates (Table~\ref{tab:audit-summary}c).
SFT improves over the base model on both groups, and the win rate on the 155 uncovered tasks (0.739, 95\% CI [0.671, 0.803]) is no lower than on the 65 covered tasks (0.692, 95\% CI [0.585, 0.800]). Together with the gains on Workspace-Bench-Lite and SpreadsheetBench~II in Table~\ref{tab:overall}, this indicates that \textbf{the improvement reflects transferable working skills rather than memorization of benchmark content}. Full details are in Appendix~\ref{app:contamination}.

\textbf{Judge sensitivity.}
We probe whether the evidence-anchored judge grounds its scores in the referenced files (Table~\ref{tab:audit-summary}d).
Deleting the worksheet that a criterion cites drops the corresponding score by 0.377 on average, and non-target criteria remain essentially unchanged (mean $|\Delta Q| = 0.016$).
This suggests that the judge reads the cited evidence and that its response is localized to the affected criterion.
In contrast, fine-grained corruptions of rows, numbers, and citations cause only small changes (below 0.03 in magnitude).
Since the corrupted cells are part of the cited evidence, an ideal judge should catch these perturbations as well. \textbf{We attribute this gap to the capability limit of GLM-5.2 as an agentic judge, which reliably detects structural evidence failures but struggles to verify fine-grained content.}

\begin{table*}[t]
\centering
\small
\caption{\textbf{Audits of the training corpus and the judge.} (a) Train--test overlap at the file, text, and occupation level. (b) Corpus construction funnel. (c) SFT transfer on occupation-covered and occupation-uncovered GDPVal tasks, with task-bootstrap confidence intervals. (d) Controlled judge perturbations.}
\label{tab:audit-summary}
\begin{minipage}[t]{0.49\textwidth}
\centering
\textbf{(a) Overlap audit}\\[2pt]
\adjustbox{max width=\linewidth}{%
\begin{tabular}{llc}
\toprule
\rowcolor{AliceBlue}
\textbf{Level} & \textbf{Comparison} & \textbf{Result} \\
\midrule
File & 39,201 train vs.\ 260 GDPVal files & 0 shared \\
Text & Top-20 13-gram pairs & 0 substantive \\
Occupation & GDPVal occupations covered & 13/44 \\
\bottomrule
\end{tabular}%
}
\end{minipage}\hfill
\begin{minipage}[t]{0.49\textwidth}
\centering
\textbf{(c) Grouped transfer, Base vs.\ SFT}\\[2pt]
\adjustbox{max width=\linewidth}{%
\begin{tabular}{lccc}
\toprule
\rowcolor{AliceBlue}
\textbf{Group} & \textbf{W/T/L} & \textbf{Win rate} & \textbf{95\% CI} \\
\midrule
Covered (65) & 43/4/18 & 0.692 & [0.585, 0.800] \\
Uncovered (155) & 109/11/35 & 0.739 & [0.671, 0.803] \\
All (220) & 152/15/53 & 0.725 & [0.668, 0.782] \\
\bottomrule
\end{tabular}%
}
\end{minipage}

\vspace{1.2em}

\begin{minipage}[t]{0.49\textwidth}
\centering
\textbf{(b) Construction funnel}\\[2pt]
\adjustbox{max width=\linewidth}{%
\begin{tabular}{lrr}
\toprule
\rowcolor{AliceBlue}
\textbf{Stage} & \textbf{Count} & \textbf{Share} \\
\midrule
Materialized tasks & 3,638 & 100.0\% \\
Unchanged rollouts reused & 2,967 & 81.6\% \\
Tasks admitted at $Q>0.90$ & 2,153 & 59.2\% \\
Validated trajectories\textsuperscript{1} & 2,169 & 59.6\% \\
Unique workspaces & 2,150 & 59.1\% \\
\bottomrule
\end{tabular}%
}
\vspace{2pt}
{\raggedright\footnotesize \textsuperscript{1}\,A task can contribute multiple trajectories when a rollout is compacted into separate training sequences.\par}
\end{minipage}\hfill
\begin{minipage}[t]{0.49\textwidth}
\centering
\textbf{(d) Controlled judge sensitivity}\\[2pt]
\adjustbox{max width=\linewidth}{%
\begin{tabular}{lr}
\toprule
\rowcolor{AliceBlue}
\textbf{Perturbation} & \textbf{Target-criterion $\Delta Q$} \\
\midrule
Deleted worksheet & $-0.377$ \\
Row corruption & $-0.018$ \\
Numeric corruption & $-0.022$ \\
Citation corruption & $-0.013$ \\
\midrule
\multicolumn{2}{c}{Non-target criteria, mean $|\Delta Q|$: $0.016$} \\
\bottomrule
\end{tabular}%
}
\end{minipage}
\end{table*}

\section{Conclusion}
\label{sec:conclusion}

We have presented GraphForge, an evidence-graph based framework that synthesizes working agent training data from real files. In GraphForge, occupational seeds fix the task direction, and an evidence graph over the instantiated workspace supplies both the task and its verification, so task requirements are backed by source files and each criterion is anchored to the files needed to verify it. Training Qwen3.6-27B on 2,169 synthesized trajectories brings GDPVal to 1445.7 (+65.7) under OpenHands, and Workspace-Bench-Lite and SpreadsheetBench~II to 63.7 (+7.7) and 24.0 (+13.7) under Claude Code, with gains holding across the OpenHands, Codex, and Claude Code scaffolds. The same data also improves Qwen3.6-35B-A3B, suggesting that GraphForge trajectories generalize across base models. Further analysis with rubric-guided rejection fine-tuning yields additional gains over SFT and supports the value of evidence-anchored selection. We hope the released data and models make it easier to build working agents that operate faithfully on real files.

\section{Discussion of Limitations}
\label{sec:limitations}
Our study has several limitations. First, our current corpus contains 2{,}169 trajectories, and we have not studied how the benefits of GraphForge scale with larger data budgets. Second, both the agent judge and the synthesis pipeline are powered by GLM-5.2. Stronger frontier models could improve the quality of the synthesized data and the reliability of the judging, and exploring the ceiling of our framework with such models remains future work. Third, our experiments cover two base models from the same family, and we do not study how GraphForge transfers to other model families. Looking ahead, we plan to scale GraphForge to more task families and file types, and to study how evidence-anchored verification interacts with longer-horizon agent scaffolds.

\newpage

\subsection*{AI use statement}

We used generative AI tools to polish the English writing and to assist with
coding tasks such as debugging. We did not use generative AI tools to design
the method, run experiments, or write the scientific claims. We reviewed all
AI-assisted content. LLM-polished text was checked by the authors for accuracy,
and LLM-generated code was verified and tested by the authors. We take full
responsibility for the final content of this work.

\subsection*{Ethics statement}

This work does not involve human subjects, so no IRB approval is required. Training workspaces are assembled from publicly available documents, such as corporate filings and public reports, and are used for research purposes only. Exact duplicates and invalid files are removed during collection. We release the trained checkpoints and the synthesized training data. The released data are built entirely from publicly available documents and are shared for research purposes under their original terms of use. Public documents may mention individuals in their original context, and we do not collect, curate, or infer any personally identifiable information beyond what already appears in these public sources. We do not foresee harmful applications of our method. We have no conflicts of interest to disclose.

\subsection*{Reproducibility statement}

The method details including architecture, training procedure, and hyperparameters are given. The training data and trained checkpoints are released at \url{https://huggingface.co/collections/groundhogLLM/graphforge}. All experiments use fixed random seeds, and the hardware and software setup is
reported.

\bibliography{iclr2027_conference}
\bibliographystyle{iclr2027_conference}

\newpage
\appendix

\section{Training Configurations}
\label{app:training-config}
Table~\ref{tab:training-config} reports the configurations used for the SFT and controlled RFT experiments. The 27B and 35B SFT models use the same corpus and optimization recipe. All three RFT arms start from the same 35B SFT checkpoint and differ only in trajectory selection.
\begin{table}[!htb]
\centering
\footnotesize
\caption{\textbf{SFT and RFT training configurations.} The anchored, unanchored, and random RFT arms use the same 462 query IDs, candidate pools, behavior filter, and optimization budget.}
\label{tab:training-config}
\begin{tabular}{@{}lll@{}}
\toprule
Configuration & SFT & RFT arms \\
\midrule
Initialization & Qwen3.6 base (27B or 35B-A3B) & 35B-A3B SFT checkpoint \\
Training examples & 2,169 trajectories & 462 matched trajectories per arm \\
Data admission & one-step revision, $Q_i>0.90$ & best-of-4, $Q_i>0.95$, behavior-clean \\
Epochs & 3 & 1 \\
Optimizer & Muon & Muon \\
Peak learning rate & $2\times10^{-5}$ & $1\times10^{-6}$ \\
Minimum learning rate & $1\times10^{-6}$ & $1\times10^{-7}$ \\
Learning-rate schedule & Cosine & Cosine \\
Warmup ratio & 0.10 & 0.10 \\
Weight decay & 0.05 & 0.05 \\
Global batch size & 8 & 8 \\
Maximum (packed) length & \multicolumn{2}{c}{262{,}144 tokens} \\
Sequence packing & Enabled & Enabled \\
\bottomrule
\end{tabular}
\end{table}
All trajectories are trained with the full assistant reasoning and tool-interaction history preserved. Samples whose complete serialized sequence exceeds 262,144 tokens are excluded rather than truncated. For RFT, the three arms use identical query IDs and training hyperparameters. Only the rule used to choose one trajectory from each four-candidate pool changes.

\section{Contamination Audit}
\label{app:contamination}

\textbf{File-level overlap.}
The 2,169 training sequences come from 2,150 unique workspaces containing
49,750 file instances and 39,201 unique SHA-256 hashes. We compared these
hashes against the files actually referenced by the 220 GDPVal tasks, which
contain 261 file instances and 260 unique hashes. No hash is shared between
the two corpora.

\textbf{Text-level overlap.}
We extracted text from every file in a supported format and added the 220 task
prompts. Extraction succeeded for 39,075 training files and 229 GDPVal
reference files. Files that failed extraction or use non-text formats still
participated in the hash check above. We retrieved candidate pairs with
normalized 13-gram signatures and manually reviewed the top 20 pairs ranked by
containment. None of them shares task requirements, entities, business facts,
or deliverable content. Five pairs share only generic numeric sequences, and
fifteen share only PowerPoint master placeholder text.

\textbf{Occupation coverage.}
13 of the 44 GDPVal occupations also appear in the training taxonomy. These
occupations account for 65 tasks, while the remaining 155 tasks belong to
occupations that the corpus does not cover. This overlap follows from the
shared O*NET taxonomy rather than from shared files or tasks.

\textbf{Grouped comparison.}
To check whether the improvement concentrates on covered occupations, we
compared the base model and the SFT model directly on all 220 tasks. Each pair
of deliverables was presented to the judge in balanced order, with the base
model shown first on 110 tasks and the SFT model shown first on the other 110.
On 212 tasks both models produced a deliverable and the judge decided the
outcome. Two tasks where only the base model produced an empty deliverable
count as wins for the SFT model, and six tasks where only the SFT model
produced an empty deliverable count as losses. Table~\ref{tab:audit-summary} reports the results. The SFT model wins at least as often on uncovered tasks as on covered ones, and the win-rate difference of 0.047 has a task-bootstrap 95\% confidence interval of $[-0.079, 0.175]$. Together with the file-level and text-level audits above, we find no sign that the improvement relies on
proximity to benchmark content.

\section{GDPVal Elo Confidence Intervals}
\label{app:gdpval-elo-ci}
Table~\ref{tab:gdpval-elo-ci} reports 95\% bootstrap confidence intervals for the GDPVal Elo scores in Table~\ref{tab:overall}. The main-table pool excludes all RFT arms. It also retains Claude Opus 4.8 (OpenHands) as a bridge node, which is needed to keep the comparison graph connected and is not reported in Table~\ref{tab:overall}. Intervals are obtained by resampling the W/T/L outcomes on each comparison edge 10{,}000 times and refitting the complete Bradley--Terry graph, with GLM-5.3 (OpenHands) fixed at 1667 in every draw.

\begin{table*}[t]
\centering
\small
\caption{\textbf{GDPVal Elo with 95\% bootstrap confidence intervals.} OpenHands and Codex results are separate (model, scaffold) nodes.}
\label{tab:gdpval-elo-ci}
\begin{tabular}{lcc}
\toprule
Model & OpenHands Elo [95\% CI] & Codex Elo [95\% CI] \\
\midrule
Claude Opus 5 & 1774.1 [1750.2, 1798.3] & 1753.1 [1697.4, 1815.7] \\
GPT-5.6-sol & 1687.1 [1663.4, 1710.7] & 1710.8 [1657.8, 1769.3] \\
Qwen3.8-Max & 1719.0 [1694.9, 1741.8] & 1771.0 [1714.6, 1831.9] \\
GLM-5.3 & 1667.0 [1667.0, 1667.0] & 1543.7 [1491.6, 1595.8] \\
Kimi-K3 & 1615.5 [1592.6, 1638.4] & 1664.4 [1613.9, 1717.2] \\
DeepSeek-V4-Pro & 1531.5 [1507.5, 1555.4] & 1576.7 [1527.0, 1628.2] \\
Nex-N2-Mini-35B & 1288.8 [1175.9, 1378.9] & 1342.3 [1296.9, 1384.5] \\
Qwen3.6-35B-A3B & 1260.6 [1195.1, 1316.3] & 1283.0 [1241.1, 1322.4] \\
\quad + SFT (GraphForge) & 1362.3 [1306.3, 1413.6] & 1384.4 [1345.8, 1422.0] \\
Qwen3.6-27B & 1380.0 [1326.6, 1432.1] & 1364.0 [1324.6, 1401.4] \\
\quad + SFT (GraphForge) & 1445.7 [1376.2, 1513.8] & 1427.4 [1389.1, 1465.2] \\
\bottomrule
\end{tabular}
\end{table*}

\section{RFT Elo Confidence Intervals}
\label{app:rft-elo-ci}
Table~\ref{tab:rft-elo-ci} reports conditional 95\% bootstrap intervals for the RFT arms in Table~\ref{tab:rft}. Each scaffold uses a star graph whose three edges compare the RFT arms directly with SFT, and the SFT point estimate from the frozen main table serves as a fixed reporting anchor. With ties counted as half wins, the Bradley--Terry solution on each edge is $400\log_{10}(p/(1-p))$ relative to SFT. For intervals, we jointly resample task UUIDs across the three edges for 10{,}000 draws and recompute the scores, so the intervals reflect task-sampling uncertainty conditional on the fixed SFT anchor. All difference intervals include zero.

\begin{table}[H]
\centering
\small
\caption{RFT direct-comparison Elo with conditional 95\% bootstrap confidence intervals.}
\label{tab:rft-elo-ci}
\begin{tabular}{lcc}
\toprule
Model & OpenHands Elo [95\% CI] & Codex Elo [95\% CI] \\
\midrule
35B SFT & 1362.3 (fixed) & 1384.4 (fixed) \\
+ RFT & 1369.5 [1319.1, 1416.5] & 1395.4 [1349.5, 1440.1] \\
+ RFT (unanchored) & 1396.4 [1348.0, 1446.3] & 1409.7 [1363.8, 1458.1] \\
+ RFT (random-of-4) & 1353.3 [1306.3, 1401.9] & 1374.9 [1330.3, 1420.8] \\
\bottomrule
\end{tabular}
\end{table}

\end{document}

%% file: math_commands.tex
\usepackage{amsmath,amsfonts,bm}

\def\eqref#1{equation~\ref{#1}}

\def\1{\bm{1}}

\DeclareMathAlphabet{\mathsfit}{\encodingdefault}{\sfdefault}{m}{sl}
\SetMathAlphabet{\mathsfit}{bold}{\encodingdefault}{\sfdefault}{bx}{n}

